# GAG-Net: Graph-Based Anatomical Guidance Network for Pharyngeal Area Segmentation in VFSS Images

JunYeong Park[1,†], Myung Woo Park[2,3,†], Byung Chan Lee[4], Hyung Jin Chang[5], Sang-Il Choi[6,*]
[1]Department of AI-based Convergence, Dankook University, Yongin, Republic of Korea
[2]Department of Rehabilitation Medicine, Seoul Metropolitan Government-Seoul National University Boramae Medical Center, Seoul, Republic of Korea
[3]Department of Rehabilitation Medicine, Seoul National University College of Medicine, Seoul National University Hospital, Seoul, Republic of Korea
[4]Department of Physical Medicine and Rehabilitation, Chung-Ang University Hospital, Seoul, Republic of Korea
[5]School of Computer Science, University of Birmingham, Birmingham, United Kingdom
[6]Department of Computer Engineering, Dankook University, Yongin, Republic of Korea
[†]These authors contributed equally to this work. [*]Corresponding author: Sang-Il Choi, choisi@dankook.ac.kr

## Abstract

Diagnosing sarcopenic dysphagia typically requires costly supplementary imaging such as computed tomography (CT) or magnetic resonance imaging (MRI) to assess muscle atrophy, burdening elderly patients. The pharyngeal area at rest ($PA_{rest}$), measurable directly from videofluoroscopic swallowing study (VFSS) images, is a promising biomarker that could eliminate this dependency, but repeated manual measurement has limited its adoption. $PA_{rest}$ segmentation is further complicated by indistinct soft-tissue boundaries and clinical rule-based delineation criteria that data-driven models capture only imprecisely. We formulate automated $PA_{rest}$ segmentation as a clinically defined measurement task, translating an implicit clinical measurement into an explicit landmark specification and annotation protocol developed with a rehabilitation medicine specialist. Applying this protocol, we constructed, to our knowledge, the first dataset for this task: 2,255 single-center VFSS frames from 137 patients, with $PA_{rest}$ contours and seven landmarks verified by a senior specialist. We then introduce GAG-Net, an automated rest-frame measurement pipeline combining deformable convolutions, for irregular pharyngeal boundaries, with a graph-based anatomical guidance module that models the geometric dependencies among landmarks prescribed by the delineation rules. On a patient-level held-out test set (237 frames, 67 patients), GAG-Net achieved the lowest area measurement error (mean absolute percentage error: 8.17% against specialist-verified references; patient-level ICC 0.97) and the highest Dice score (0.8862) among all evaluated methods, with 67.2% fewer parameters and 74% less training memory than the two-stage JOINED model. Integrating clinician-defined anatomical knowledge may enable automated $PA_{rest}$ measurement from VFSS images alone, supporting quantitative assessment of sarcopenic dysphagia without supplementary imaging.

## 1. Introduction

Sarcopenic dysphagia is a swallowing disorder that develops when sarcopenia — muscle loss and functional decline associated with the frailty process — affects the swallowing musculature [1]. It is directly tied to survival [2, 3]: one in three patients with aspiration pneumonia, the most severe complication of dysphagia, has coexisting sarcopenia [4]. Despite this severity, a definitive diagnosis remains difficult to obtain.

Videofluoroscopic Swallowing Study (VFSS) is the gold standard for diagnosing dysphagia [5, 6, 7], but it cannot confirm sarcopenic dysphagia on its own. A definitive diagnosis additionally requires evidence of muscle atrophy, typically obtained through computed tomography (CT) or magnetic resonance imaging (MRI) [1, 8]. For elderly patients with limited mobility, these additional scans add appointments and waiting time to an already burdensome diagnostic process.

This study aims to identify a quantitative anatomical marker for muscle atrophy that VFSS alone can provide. We focus on the pharyngeal area at rest ($PA_{rest}$), the pharyngeal airspace on a pre-swallow rest frame, bounded by seven anatomical landmarks rather than a visually distinct edge [9]. $PA_{rest}$ measurement is significantly associated with sarcopenic dysphagia [10], and CT-based volumetry confirms it reflects actual pharyngeal muscle atrophy [11], indicating that $PA_{rest}$ is a validated VFSS-derived biomarker. However, $PA_{rest}$ measurement currently relies on manual segmentation, which is time-consuming and rater-dependent, limiting its use in routine clinical practice [11]. Automating this process is therefore essential to making $PA_{rest}$ clinically usable.

Deep learning has advanced automated VFSS analysis — from airway invasion classification and hyoid bone tracking [12, 13] to temporal segmentation models [14, 15, 16]. These methods, however, predominantly target structures with distinct visual boundaries. Consequently, they are ill-suited for $PA_{rest}$ segmentation. Automated measurement of clinical indices defined by anatomical landmarks has precedent in imaging informatics, for example the Insall–Salvati ratio derived from keypoints detected on lateral knee radiographs [17]. Such an index reduces to distances between landmarks, whereas $PA_{rest}$ is a region area bounded partly by landmark-defined lines.

$PA_{rest}$ boundaries are blurred where they cross soft tissue. The posterior pharyngeal wall, tongue base, and pharyngeal constrictor muscles lack distinct radiographic contrast in VFSS, and the landmarks needed to localize these boundaries — the nasal spine, atlas, and arytenoid cartilage, for instance — are themselves frequently obscured or overlapped by adjacent structures in a single 2D frame. A more fundamental obstacle, however, is that clinical protocols define several boundary segments not by tissue appearance at all, but by geometric rule — a straight line connecting two reference landmarks, for instance [9, 10, 18, 19]. Such rules encode relational, symbolic constraints between landmarks rather than local visual patterns. Appearance-based segmentation models can approximate these constraints through learned priors on head and neck geometry, but such approximation lacks the precision that explicit landmark grounding provides. $PA_{rest}$ segmentation therefore requires landmark localization and anatomical reasoning in addition to conventional boundary delineation.

To address these challenges, we first translate the clinical delineation criteria into an explicit landmark specification and annotation protocol through consultation with a rehabilitation medicine specialist. Applying this protocol, we construct a dataset of 2,255 VFSS frames from 137 patients, with $PA_{rest}$ contours and seven anatomical landmarks verified by a senior specialist.

Building on this formulation, we present the Graph-based Anatomical Guidance Network (GAG-Net), an automated $PA_{rest}$ measurement pipeline for rest frames, built on a U-Net architecture [20]. GAG-Net pairs two mechanisms, each targeting one of the obstacles above: Deformable Convolutional Networks (DCN) in the encoder adaptively sample around irregular, low-contrast soft-tissue boundaries, while a Graph-based Anatomical Guidance Module (GAGM) encodes the relational structure among anatomical landmarks and passes it to the decoder as guidance. Unlike prior landmark-guided methods relying on pairwise distance constraints [21, 22], this graph-based formulation and its single end-to-end pipeline together achieve accurate segmentation and area measurement with substantially fewer parameters, requiring no manual landmark or contour input at inference.

These results position GAG-Net as a viable path toward moving $PA_{rest}$ measurement out of manual annotation and into routine clinical use. The contributions of this study are threefold.

1. We are the first, to our knowledge, to formulate automated $PA_{rest}$ segmentation as a well-defined, reproducible measurement task: through iterative consultation with a rehabilitation medicine specialist, we translated an implicit, manually-performed clinical measurement into an explicit landmark specification and annotation protocol amenable to deep learning. Building on this formulation, we present an end-to-end framework that reduces the repeated manual labor required for $PA_{rest}$ measurement, addressing a key barrier that has kept this validated biomarker out of routine clinical use.

2. Applying this protocol, we construct, to our knowledge, the first dataset for this task: 2,255 VFSS frames from 137 patients with $PA_{rest}$ contours and seven anatomical landmarks, annotated by a trained annotator and verified by a senior rehabilitation medicine specialist — distinguishing it from prior VFSS datasets built around directly visible structures.

3. We show that explicitly encoding anatomical landmark structure through a graph-based guidance module improves segmentation accuracy and area measurement precision over both conventional CNN baselines and vision foundation models, while using substantially fewer parameters.

## 1.1 Related Work

### *1.1.1 Deep Learning for VFSS Analysis*

Deep learning applications in VFSS have primarily concentrated on temporal phase classification, anatomical tracking, and anomaly detection. Early approaches utilized optical flow to identify motion-intensive segments for automated pharyngeal phase detection [23]. Subsequent work applied deep learning to increasingly specific structures — Lee et al. [12] for frame-level airway invasion classification via cervical spine segmentation, others for hyoid bone tracking [13], bolus segmentation [16], and multi-structure segmentation via knowledge distillation [24].

These methods succeed precisely because their targets — the airway, the hyoid bone, and the bolus — present distinct radiographic contrast that a convolutional network can learn to trace. $PA_{rest}$ offers no such target. Its boundary is defined by clinical rule rather than tissue contrast, placing it outside the scope of every approach above [25, 26].

Recent work has extended VFSS analysis toward quantitative swallowing assessment. Yuan et al. [27] segment the pharyngeal cavity, laryngeal vestibule, and bolus together with five anatomical keypoints, and derive frame-wise area curves and kinematic parameters over swallow segments. Their pharyngeal cavity is a functional region tracked over time. $PA_{rest}$ differs in that it is a single measurement on a rest frame, its contour is defined by clinical rules between anatomical landmarks, and it serves as a biomarker for sarcopenic dysphagia.

### *1.1.2 Anatomical Priors in Medical Image Segmentation*

Incorporating anatomical priors is a well-established strategy to enhance the clinical validity of segmentation. Liu et al. [28] systematically categorized anatomy-aided deep learning methods into shape, appearance, motion, and context priors, and identified graph neural network-based structural reasoning as a promising but underexplored direction. Traditional methods integrated global shape priors, such as statistical atlases or mean shapes, to constrain predictions within anatomically plausible manifolds [29, 30]. However, these rigid priors often fail to accommodate significant inter-patient variability or complex non-rigid deformations. Alternative strategies have explored attention-based [31] and cross-modal [32] guidance to handle boundary ambiguity in medical image segmentation, though without explicitly modeling anatomical landmark relationships.

Recent studies have shifted toward more flexible priors based on anatomical landmarks. Joham et al. [33] modeled spatial constraints using differentiable Markov Random Fields to refine landmark predictions, while others [21, 22] incorporated pixel-wise landmark distance regression via multi-task learning. Because these methods constrain each landmark pair independently, an error in one predicted landmark neither propagates to nor benefits from its neighbors. A graph-based formulation removes this independence, representing landmarks as nodes so that a single uncertain prediction is constrained by the joint configuration of all landmarks through message passing, rather than by any one pairwise distance. Larrazabal et al. [34] adopted denoising autoencoders as a post-processing step to enforce anatomical plausibility, but such reactive approaches cannot influence the segmentation during inference. More broadly, existing methods treat landmarks as auxiliary targets or apply anatomical constraints only during post-processing, rather than as an intrinsic part of the inference pipeline itself.

Landmark-based automation of clinical measurements has also been reported in imaging informatics. Adleberg et al. [17] trained a keypoint detection model to measure the Insall–Salvati ratio on lateral knee radiographs and evaluated it with landmark distance error, correlation against manual labels, and Bland–Altman analysis on internal and external validation sets. In that setting, the index follows from distances between detected keypoints. $PA_{rest}$ differs in that the measured quantity is a region area, so the contour itself must be delineated, including segments defined as straight lines between landmarks, and localizing the landmarks alone does not determine the measurement.

#### *1.1.3 Graph-based and Deformable Convolution Architectures*

Graph Convolutional Networks (GCNs) and Deformable Convolutional Networks (DCNs) have emerged as complementary tools. Tian et al. [35] proposed ARR-GCN, which explicitly learns inter-region anatomical relations via graph reasoning for fine-grained organ segmentation in CT. Similarly, Liang et al. [36] proposed CylinGCN, a contour-based graph convolutional network that encodes cylindrical topological priors for continuous 3D segmentation of tubular anatomical structures such as airways and vessels, addressing the discontinuity artifacts of classification-based approaches. Gaggion et al. [37] demonstrated that hybrid graph neural networks improve anatomical plausibility in chest X-ray segmentation by encoding landmark connectivity through graph adjacency matrices. Li et al. [38] further extended GCN-based structural reasoning to the temporal domain, modeling spatio-temporal landmark dependencies in echocardiographic sequences. However, these methods have primarily been applied to structures with relatively well-defined visual boundaries (e.g., liver segments, lung lobes, cardiac chambers). Whether they extend to targets whose boundaries are clinically defined rather than visually apparent has not been tested.

Complementing GCNs, DCNs address local geometric variations by dynamically adjusting sampling locations based on learned offsets [39]. Hu et al. [40] demonstrated the effectiveness of adaptive convolution strategies for medical anatomy segmentation, proposing dynamic strip convolutions tailored for orientation-varying structures. While DCNs enhance local boundary adaptability, they lack the capacity to enforce global structural constraints. GCNs and DCNs are therefore complementary — GCNs for global structural reasoning, DCNs for local boundary adaptation. Yet no prior work combines them for tasks, such as $PA_{rest}$ segmentation, that require both capabilities simultaneously.

# 2. Materials and Methods

## 2.1 Data Collection and Annotation

### *2.1.1 Dataset*

The dataset used in this study consists of retrospectively collected VFSS X-ray videos obtained from the Department of Rehabilitation Medicine at Chung-Ang University Hospital. A total of 2,383 VFSS studies from patients evaluated for dysphagia between January 2023 and July 2024 were initially screened under Institutional Review Board (IRB) approval (IRB No. 2501-007-19557). From these, 248 videos from 137 unique patients were selected; the remaining studies were excluded due to insufficient image quality, absence of a clearly identifiable rest frame, or incomplete anatomical landmark visibility. All selected frames were anonymized by uniformly cropping peripheral regions to remove patient identifiers. The videos were decomposed into individual frames, and only those clearly depicting the $PA_{rest}$ were selected through a structured data refinement process. Frames depicting $PA_{rest}$ were selected based on specialist-validated inclusion criteria:

(1) complete contact between the soft palate and the tongue base; (2) absence of severe mandible misalignment; (3) residue obscuring <50% of the epiglottis or arytenoid cartilage; (4) clear visibility of the key anatomical landmarks.

### *2.1.2 VFSS Protocol*

The VFSS procedure was performed following the protocol described by [7], with minor modifications. Videofluoroscopic images were acquired via lateral projection and stored digitally at a speed of 60 fps (frames per second) while the patient swallowed boluses of various consistencies mixed with contrast medium in a seated position. Materials of different consistencies were administered using the International Dysphagia Diet Standardisation Initiative (IDDSI) framework [41] (Levels 0–7, where Level 0 indicates thin liquids and higher levels indicate thicker liquids or more textured solids). The examination began with an initial 2 mL thin-liquid trial (IDDSI Level 0; water). Subsequent bolus trials were selected as clinically indicated based on patient tolerance and safety, and could include 5 mL thin liquid (IDDSI Level 0; water), slightly thick liquid (IDDSI Level 1; juice-consistency), mildly thick liquid (IDDSI Level 2; yogurt-consistency), a regular-texture solid bolus (IDDSI Level 7; Regular), and porridge with small soft particles (approximately IDDSI Level 5; Minced & Moist). When deemed safe, self-administered cup drinking of thin liquid (IDDSI Level 0; water, ≥ 10 mL) was additionally performed. The study was suspended immediately when aspiration was evident, under the same IRB approval described in Section 2.1.1.

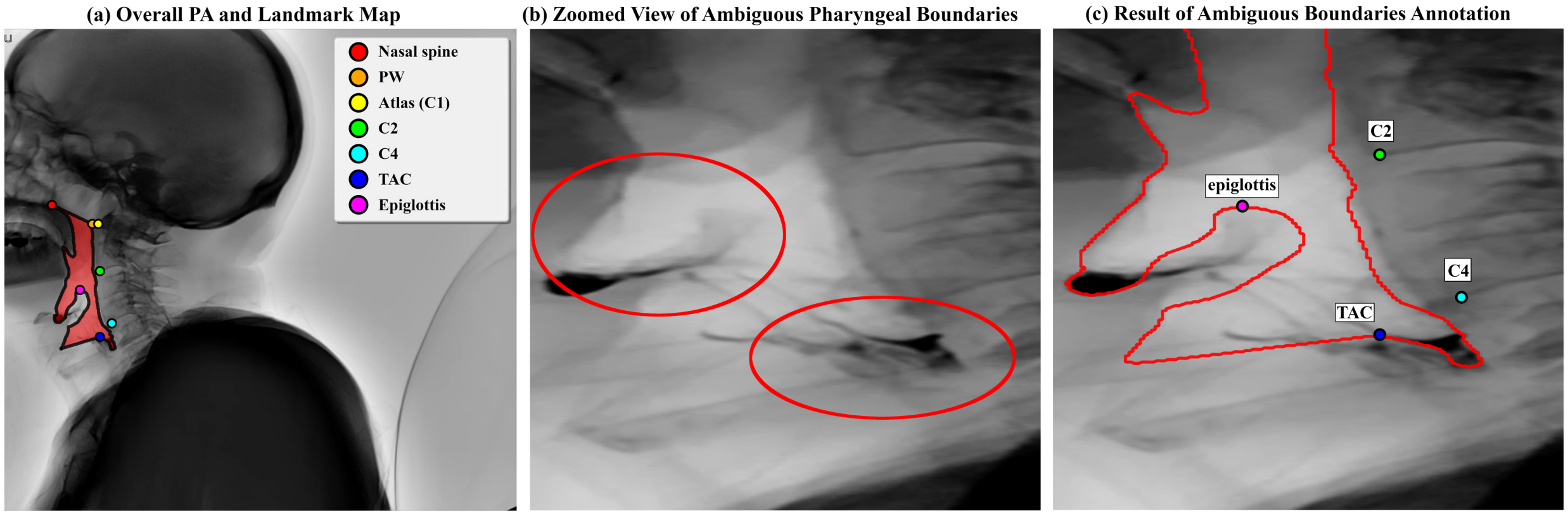

**Fig. 1** Definition and annotation challenges of the $PA_{rest}$. (a) Delineation of the $PA_{rest}$ (outlined polygon) and seven key anatomical landmarks (dots; see legend). (b) Zoomed-in view highlighting boundary ambiguity, where low contrast and overlapping soft tissues obscure exact edges (ellipses). (c) Ground-truth annotation of the same region. Despite visual ambiguity, boundaries are defined by leveraging the spatial context of adjacent landmarks (e.g., Epiglottis, C2, TAC), demonstrating the necessity of anatomical priors

### *2.1.3 Annotation Protocol and Landmark Specification*

Fig. 1 presents the annotated $PA_{rest}$ region along with the seven key anatomical landmarks. All $PA_{rest}$ annotations were performed by a single trained annotator following a standardized training procedure, and subsequently verified by a senior Rehabilitation Medicine specialist (>10 years of experience) to ensure clinical accuracy and consistency. The $PA_{rest}$ contour was delineated in strict accordance with criteria established in prior studies [10, 18, 19]. The $PA_{rest}$ boundaries were defined as follows:

**Superior boundary**: A straight line connecting the posterior nasal spine toward the midpoint of the tubercle of the Atlas (C1); this line terminates where it meets the posterior pharyngeal wall, a point annotated as the beginning of the Posterior Wall (PW).

**Posterior boundary**: From PW, the contour follows the posterior pharyngeal wall downward to the level of the superior border of the arytenoid cartilage.

**Anterior and inferior boundary:** Following the pyriform sinus contour to the top of the arytenoid cartilage (TAC), a straight line perpendicular to the C2–C4 line extends anteriorly to the anterior pharyngeal wall. From there, the boundary traces the epiglottis and tongue base superiorly, returning to the nasal spine to complete the region.

In practice, VFSS images often contain anatomical deformation, structural overlap, or contrast-media artifacts, which can obscure boundary visibility. To address these issues and maintain annotation consistency, we defined additional detailed rules in consultation:

- The pyriform sinus, located posterior to the arytenoid cartilage, is included within the $PA_{rest}$ region.
- The pyramidal structure superior to the cricoid cartilage is defined as the arytenoid cartilage.
- For the epiglottis, whose morphology can be difficult to distinguish, the entire visible silhouette is included to ensure consistency.
- When residue above the epiglottis cannot be captured with a single contour due to alignment issues, it is annotated separately.

These rule-based criteria operationalize implicit clinical judgment into reproducible annotation decisions. To support anatomically grounded $PA_{rest}$ boundary estimation, we also defined seven key anatomical landmarks, namely nasal spine, beginning of Posterior Wall (PW), atlas, C2, C4, Top of Arytenoid Cartilage (TAC), and epiglottis, based on the criteria described in [10]. Each landmark was stored as an ($x$, $y$) coordinate in an XML file and subsequently used as structural guidance input for the proposed graph-based anatomical module.

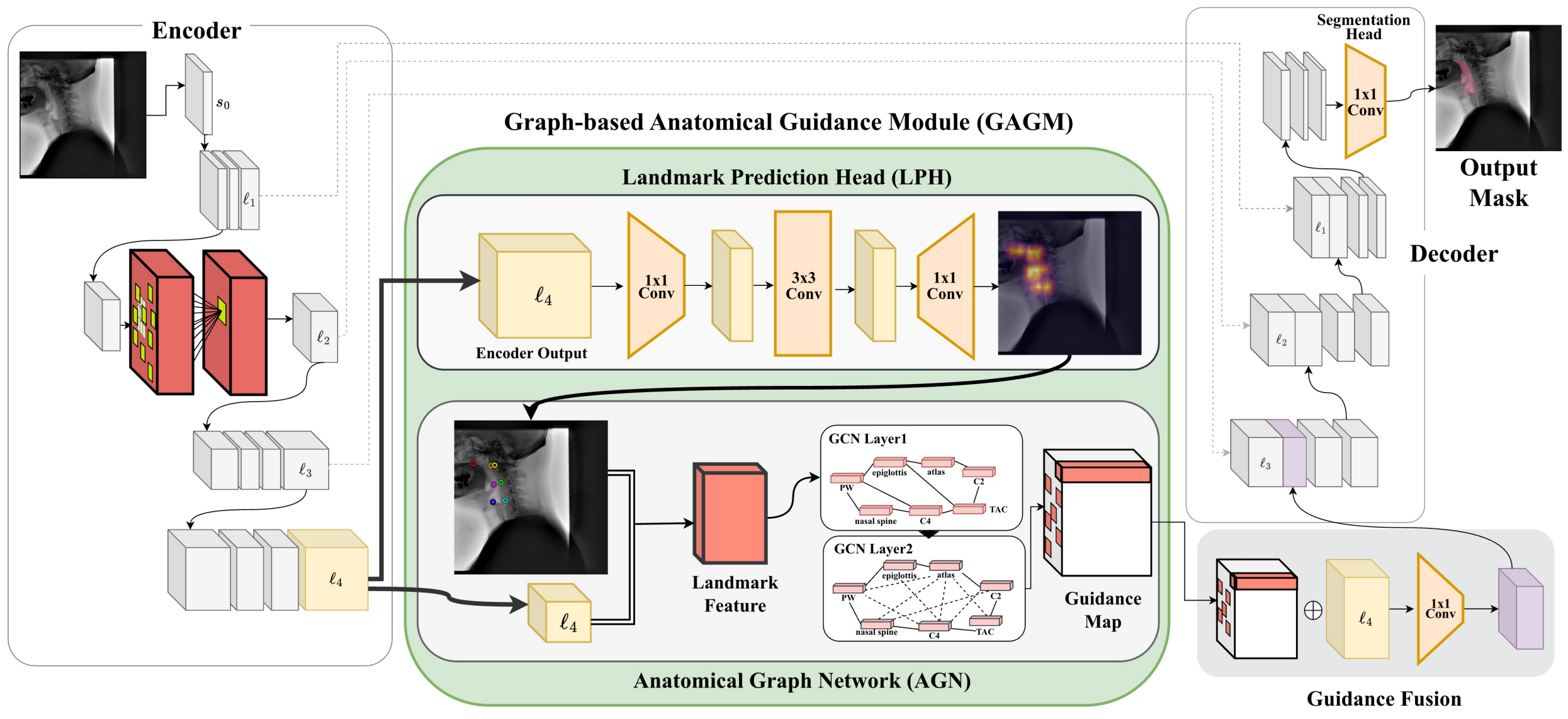

**Fig. 2** Overall architecture of GAG-Net. The framework comprises three primary modules: (1) Encoder: A modified ResNet-50 backbone in which deformable convolutions (highlighted blocks) are integrated into the second stage to robustly handle anatomical shape variations. (2) Graph-based Anatomical Guidance Module (GAGM): A Landmark Prediction Head (LPH) estimates one heatmap per landmark from deep encoder features ($l_4$), and the landmark coordinates obtained from the heatmaps are processed by the Anatomical Graph Network (AGN) to generate a structural Guidance Map. The heatmaps and coordinates shown are model outputs for an example test frame, and the internal structure of the AGN is shown in Fig. 4. (3) Decoder: Generates the final segmentation mask by fusing the anatomical Guidance Map with multi-scale encoder features via skip connections and upsampling

## 2.2 Graph-based Anatomical Guidance Network (GAG-Net)

$PA_{rest}$ boundaries are inherently ambiguous due to low radiographic contrast between adjacent soft tissues and substantial inter-patient anatomical variation (Fig. 1(b),(c)). Motivated by the challenges established in Section 1, we propose GAG-Net, an automated $PA_{rest}$ measurement pipeline for rest frames that augments CNN-based segmentation with explicit anatomical priors through two complementary mechanisms: landmark localization via a dedicated detection head, and anatomical constraint enforcement via graph-based spatial reasoning.

GAG-Net builds upon a U-Net [20] architecture (Fig. 2). In the encoder, we integrate deformable convolution layers alongside residual convolutions to better capture fine boundary characteristics and accommodate the considerable morphological variability of the pharyngeal region. Furthermore, to incorporate anatomical landmark information beyond pixel-level features, we introduce a Graph-based Anatomical Guidance Module (GAGM) between the encoder and decoder, enabling the model to leverage latent structural information present in VFSS images.

### *2.2.1 Preliminary*

U-Net [20] is a widely used encoder-decoder architecture for medical image segmentation. The encoder progressively reduces spatial resolution through convolution and pooling to extract high-level representations, while the decoder restores spatial detail using upsampling and skip connections. This symmetric design enables the joint capture of global context and fine-grained structure.

Deformable Convolutional Networks (DCNs) [39] extend standard convolution by introducing learnable offsets to the fixed sampling grid, allowing sampling locations to adapt to local geometric changes. This flexibility enables the model to capture regions with substantial shape variation or indistinct boundaries, making DCNs well-suited for modeling the irregular contours of the $PA_{rest}$.

Graph Convolutional Networks (GCNs) [42] operate on graph-structured data by aggregating features from neighboring nodes, thereby encoding both local and global structural dependencies. In GAG-Net, anatomical landmarks are represented as nodes and their spatial relationships as edges, enabling the model to learn the structural constraints that define the $PA_{rest}$.

### *2.2.2 Encoder Configuration with Deformable Convolution*

The encoder backbone is a ResNet-50 [43] pretrained on ImageNet [44], producing multi-scale features at 1/2, 1/4, 1/8, 1/16, and 1/32 of the input resolution (512 × 512). To handle the irregular and non-rigid boundaries of the $PA_{rest}$, we replace the 3 × 3 convolutions in all four bottleneck blocks of layer2 with Deformable Convolution (DCN) [39] (highlighted blocks in Fig. 2). Each DCN module consists of an offset prediction layer that generates 2 × 3 × 3 = 18 offset values, followed by the deformable convolution itself. The offset prediction layer is initialized to zero, ensuring that the deformable convolution begins as a standard convolution and gradually learns adaptive sampling locations during training. To further stabilize early training, DCN offset parameters are frozen for the first 10 epochs and subsequently unfrozen for joint optimization.

This adaptive sampling capability is particularly suited to $PA_{rest}$, whose boundaries are shaped by both passive anatomical structures and active physiological processes such as bolus flow, yielding irregular, patient-specific contours that fixed sampling grids cannot follow (Fig. 3).

Residual connections mitigate gradient vanishing and stabilize training by combining deformable and standard convolutional features, providing robust representations for the subsequent landmark-based graph reasoning and decoder-stage segmentation.

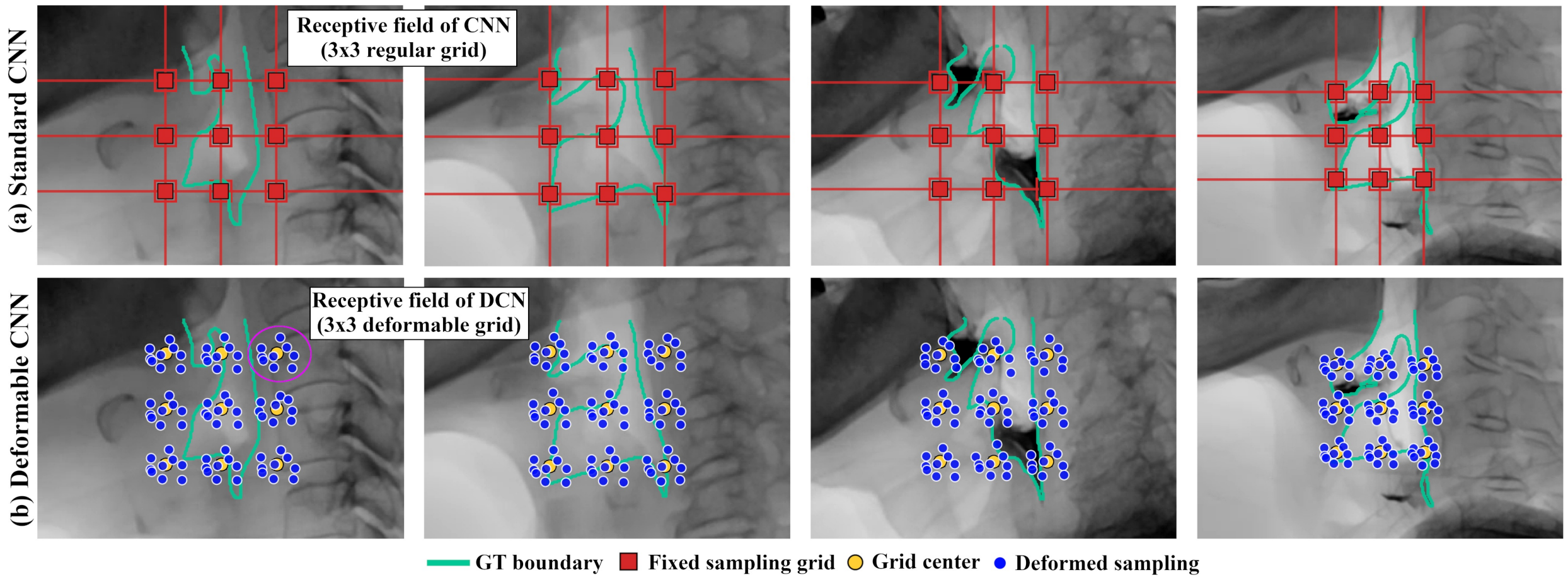

**Fig. 3** Comparison of sampling mechanisms between standard and deformable convolution. (a) Standard Convolution: Features are sampled from a fixed, rigid 3× 3 grid (squares), which limits the ability to model irregular structures. (b) Deformable Convolution: Sampling locations are dynamically adjusted. Adaptive offsets are calculated relative to the grid center (circle), causing the actual sampling points (dots) to shift and actively align with the irregular anatomical boundaries (contour line). This contrasts with the rigid sampling of standard convolution

### *2.2.3 Graph-based Anatomical Guidance Module (GAGM)*

GAGM comprises two primary components: the Landmark Prediction Head (LPH) and the Anatomical Graph Network (AGN) (Fig. 2). Adopting a multi-task joint learning strategy [21], GAGM uses landmark prediction as auxiliary supervision to resolve ambiguous $PA_{rest}$ boundaries. Unlike prior methods relying on simple pairwise distances [21, 22] between anatomical landmarks, GAGM explicitly models complex global interdependencies via graph representations. Notably, this design enhances structural coherence while requiring only 32.8% of JOINED's combined parameter count (39.7M vs. 121.2M across both stages), and does so within a unified end-to-end training framework, a key advantage over prior approaches that required separate two training stages.

Given this architectural relationship, JOINED [21] was selected as a structural baseline for the comparison in Section 3.3: although originally designed for optic disc/cup segmentation, its target domain shares key characteristics with $PA_{rest}$, namely visually indistinct boundaries that are difficult to delineate manually and non-rigid morphological variability. Code for other landmark-guided methods in medical imaging was not publicly available, so JOINED was the only such method we could compare against.

***1) Landmark Prediction Head*** The LPH consists of three convolutional layers applied to $l_4 \in \mathbb{R}^{2048 \times 16 \times 16}$: a 1 × 1 convolution reducing channels from 2048 to 256, followed by a 3 × 3 convolution maintaining 256 channels, and a final 1 × 1 convolution producing 7 output channels (one per landmark). Each of the first two layers is followed by batch normalization and ReLU activation. The resulting 16 × 16 heatmaps are bilinearly upsampled to the original input resolution (512 × 512) for loss computation. The predicted heatmaps (one channel per landmark) and the landmark coordinates obtained from them by soft-argmax are shown for an example test frame in Fig. 2.

To ensure training stability, we employ a scheduled sampling strategy [45] with a threshold epoch $\tau = 20$:

- **Teacher Forcing Phase (epoch < τ):** The AGN receives ground-truth landmark coordinates as input, which are independent of the LPH output. Consequently, no gradients from the AGN flow back to the LPH during this phase, allowing the graph module to learn stable anatomical dependencies without being affected by early, unreliable landmark predictions. The model is optimized using both the segmentation loss ($L_{seg}$) and the landmark prediction loss ($L_{lmk}$).
- **Self-Guided Phase (epoch ≥ τ):** Once landmark prediction has stabilized, the AGN transitions to using coordinates derived from the LPH output via soft-argmax. This enables the full pipeline to operate end-to-end during inference without requiring ground-truth annotations.

***2) Anatomical Graph Network (AGN)*** The AGN leverages Graph Convolutional Networks (GCNs) [42] to encode global anatomical topology. We construct a graph where the seven landmarks form nodes connected by a chain adjacency structure with self-loops (i.e., each landmark is connected to its immediate neighbors in the anatomical sequence: nasal spine ↔ PW ↔ atlas ↔ C2 ↔ C4 ↔ TAC ↔ epiglottis). The adjacency matrix is fixed and not learned. Through message passing, the AGN aggregates structural information to generate a Guidance Map ($G_{Map}$) that encodes spatial constraints within the $PA_{rest}$ region. This map is fused with encoder features, ensuring the decoder receives anatomically coherent feature representations that integrate both local visual cues and global anatomical priors.

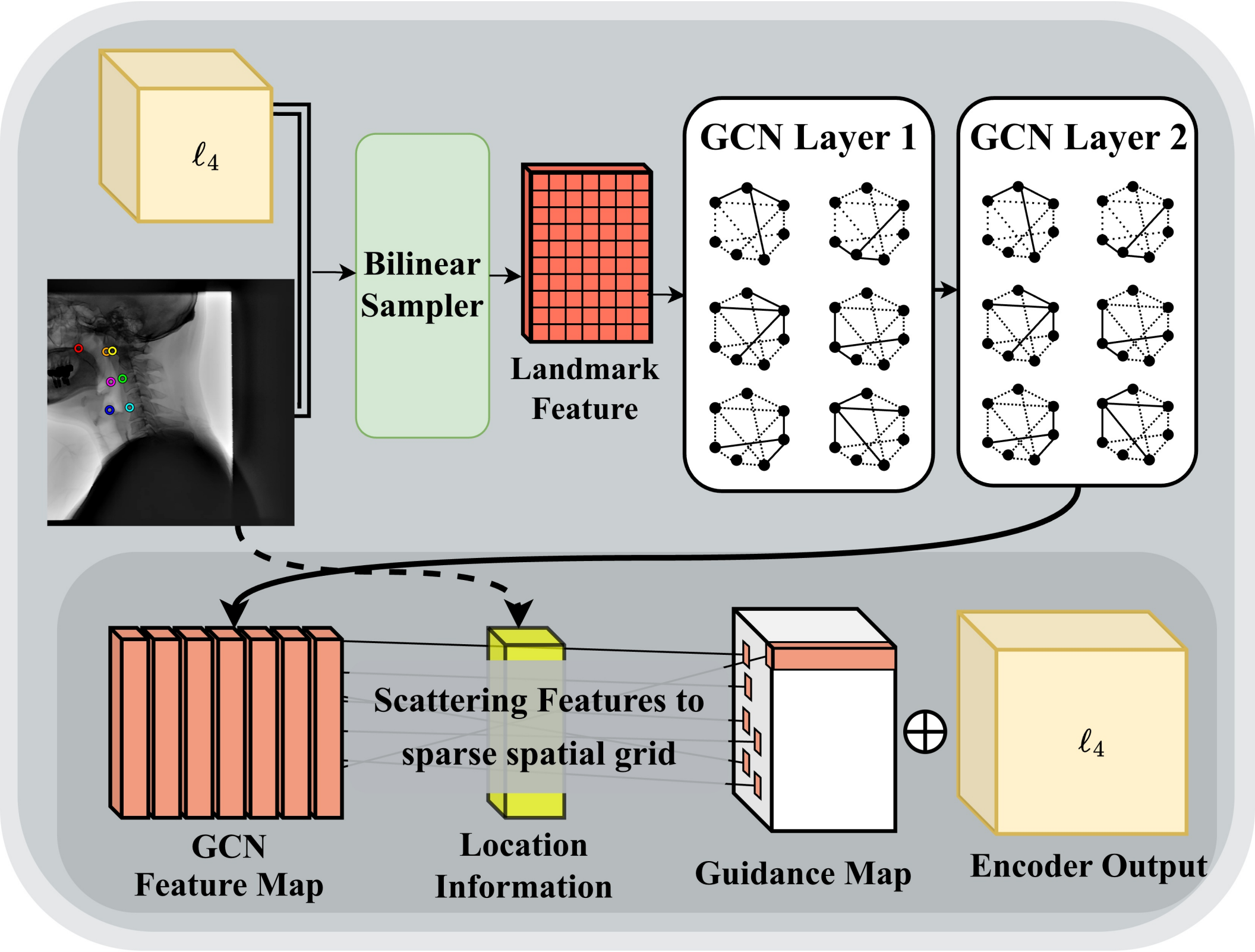

**Fig. 4** Detailed operational pipeline of the Anatomical Graph Network (AGN). (1) Landmark features are sampled from the encoder output ($l_4$) via bilinear interpolation using coordinate inputs. (2) Two sequential GCN layers process these features to encode global structural dependencies. (3) The refined features are scattered onto a sparse spatial grid to form the Guidance Map ($G_{\text{Map}}$), which is subsequently fused with $l_4$ via concatenation and convolution to provide anatomically enriched features for the decoder

***Operational Pipeline of AGN*** Fig. 4 details the workflow of the Anatomical Graph Network (AGN). The AGN processes high-level encoder features and landmark coordinates to generate a structural Guidance Map ($G_{\text{Map}}$) through three sequential stages:

1. **Feature Sampling and Input Strategy:**
   The module inputs are the final encoder feature map $l_4 \in \mathbb{R}^{2048 \times H/32 \times W/32}$ and landmark coordinates $P \in \mathbb{R}^{L \times 2}$ (where $L$=7). Feature vectors corresponding to each landmark are extracted from $l_4$ via bilinear interpolation, resulting in a landmark feature matrix $F_{\text{lmk}} \in \mathbb{R}^{L \times 2048}$. To ensure robust training, we adopt a scheduled sampling strategy: (1) *Teacher Forcing Phase* (epoch ≤ τ) uses ground-truth coordinates to stabilize initial graph learning; (2) *Self-Guided phase* (epoch > τ) uses predicted coordinates from the LPH, establishing end-to-end optimization.

2. **Graph Reasoning via GCN:**
   The landmark features $F_{\text{lmk}}$ serve as nodes in a graph where edges encode anatomical relationships. Two sequential GCN layers [42] aggregate and update node features from 2048 to 256 dimensions ($h_G$=256), with ReLU activation applied after the first layer. The refined features are denoted $F_{\text{refine}} \in \mathbb{R}^{L \times 256}$.

3. **Guidance Map Generation and Feature Fusion:**
   The refined node features $F_{\text{refine}} \in \mathbb{R}^{7 \times 256}$ are scattered onto a sparse 16 × 16 spatial grid at the corresponding landmark coordinates, producing $G_{\text{Map}} \in \mathbb{R}^{256 \times 16 \times 16}$. Although this map is sparse (7 non-zero locations out of 256 spatial positions), the subsequent concatenation with the dense encoder feature $l_4$ and fusion via a 1 × 1 convolution (2048 + 256 → 512) ensures that the anatomical signals are propagated through the decoder alongside the full visual feature representation.

#### *2.2.4 Decoder and Segmentation Head*

The decoder consists of five upsampling stages. At each stage, the feature map is bilinearly upsampled by a factor of 2 and concatenated with the corresponding encoder skip connection, followed by two 3 × 3 convolution layers with batch normalization and ReLU. The channel progression through the decoder is: 512 → 512 (skip from $l_3$, 1024 channels) → 256 (skip from $l_2$, 512 channels) → 128 (skip from $l_1$, 256 channels) → 64 (skip from $s_0$, 64 channels) → 32 (no skip). A final 1 × 1 convolution with sigmoid activation produces the binary segmentation mask.

#### *2.2.5 Loss Function*

GAG-Net is optimized end-to-end using:

$$\mathscr{L}_{\text{total}} = \mathscr{L}_{\text{seg}} + \lambda_{\text{lmk}} \cdot \mathscr{L}_{\text{lmk}}, \tag{1}$$

where $L_{\text{seg}} = L_{\text{Dice}} + L_{\text{BCE}}$ combines Dice loss and binary cross-entropy loss with equal weighting to balance overlap-based and pixel-wise accuracy. $L_{\text{lmk}}$ is the mean squared error between predicted and ground-truth landmark heatmaps, weighted by $\lambda_{\text{lmk}} = 1.0$ throughout training.

### 2.3 Implementation Details

All experiments were implemented in PyTorch 2.8 on a single NVIDIA A100 GPU with 80GB of VRAM. GAG-Net utilizes a U-Net architecture with a ResNet-50 encoder, where deformable convolution layers are integrated into the second residual stage to accommodate irregular pharyngeal boundaries.

**Data Splits and Patient-Level Organization.** For model development, the 2,255 annotated frames were split into training, validation, and test sets at the patient level to minimize subject overlap across splits. The training, validation, and test sets comprised 1,916 frames from 59 patients, 102 frames from 11 patients, and 237 frames from 67 patients, respectively, totaling 2,255 frames from 137 patients. Across the entire dataset, the number of annotated frames per patient ranged from 1 to 154. To enable robust feature learning during training, the training set was composed of patients with higher frame density per subject (mean 32.5 frames), allowing the model to capture patient-specific anatomical patterns. Conversely, to ensure fair generalization assessment and mitigate bias from patient-specific characteristics, the test set was deliberately constructed with a larger patient cohort (67 patients) but lower per-patient frame density (mean 3.5 frames, range 1–9). This design prevents spurious high performance arising from memorizing individual patient anatomy rather than learning generalizable segmentation patterns.

**Optimization and Training Strategy.** The model was optimized using AdamW with an initial learning rate of $2 \times 10^{-4}$ and weight decay of $1 \times 10^{-4}$. We employed a ReduceLROnPlateau scheduler [46], decaying the learning rate by a factor of 0.3 after 8 epochs of validation stagnation. Training spanned 150 epochs with a batch size of 16. To prevent overfitting, data augmentation (horizontal flip, random rotation ± 10°) was applied after the first 25 epochs.

**Progressive Anatomical Guidance.** We adopted a progressive strategy to stabilize the training of the graph module. For the first 20 epochs (Teacher Forcing phase), the AGN was supervised with ground-truth landmarks to learn reliable anatomical dependencies. Subsequently (Self-Guided phase), the model transitioned to using landmarks predicted by the LPH, aligning the training process with inference conditions and enhancing robustness to prediction uncertainty.

**Resolution and Baselines.** GAG-Net and standard CNN baselines were trained at 512 × 512 resolution. For fair comparison, foundation models (SAM [47], MedSAM [48]) were evaluated at their native 1024 × 1024 resolution using automatically generated bounding-box prompts. The fine-tuned SAMUS [49] was initialized with MedSAM weights.

## 2.4 Evaluation Metrics

Segmentation performance was quantified using metrics that assess volumetric overlap, boundary fidelity, and clinically relevant area measurement precision.

### *2.4.1 Segmentation Overlap Metrics*

**Dice Similarity Coefficient (DSC).** The DSC measures the volumetric overlap between the predicted mask $P$ and ground truth $G$:

$$DSC = \frac{2 \times |P \cap G|}{|P| + |G|} \tag{2}$$

where |·| denotes the cardinality of foreground pixels, ranging from 0 to 1 with 1 indicating perfect overlap.

**Jaccard Index (IoU).** The Jaccard index quantifies overlap between two sets:

$$\text{Jaccard} = \frac{|P \cap G|}{|P \cup G|} = \frac{|P \cap G|}{|P| + |G| - |P \cap G|} \tag{3}$$

Like DSC, Jaccard ranges from 0 to 1, but penalizes false positives more heavily.

### *2.4.2 Boundary Distance Metrics*

**95th Percentile Hausdorff Distance (HD95).** HD95 evaluates boundary adherence by measuring the 95th percentile distance between the contours of predicted and ground truth masks, reducing sensitivity to outliers:

$$HD95 = \text{percentile}_{95}\left(\{d(p, \partial G) : p \in \partial P\} \cup \{d(g, \partial P) : g \in \partial G\}\right) \tag{4}$$

where $\partial P$ and $\partial G$ represent the boundaries of predicted and ground truth masks, and $d(\cdot, \cdot)$ denotes the minimum Euclidean distance. All distance metrics were computed in pixel units with a spacing parameter of (1.0, 1.0) to maintain consistency with original evaluation pipelines.

**Average Symmetric Surface Distance (ASD).** The ASD measures the average distance between the surfaces of predicted and ground truth masks:

$$ASD = \frac{1}{|\partial P| + |\partial G|}\left(\sum_{p \in \partial P} d(p, \partial G) + \sum_{g \in \partial G} d(g, \partial P)\right) \tag{5}$$

Unlike HD95, which captures the worst-case boundary error via the 95th percentile, ASD provides a mean-based measure sensitive to systematic boundary offset across the entire contour.

### *2.4.3 Clinical Area Measurement Metrics*

**Pharyngeal Area Calculation.** The $PA_{rest}$ was computed from binary segmentation masks as:

$$A_{PA_{rest}} = N_{\text{pixels}} \times (\text{mm_per_pixel})^2 \tag{6}$$

where $N_{pixels}$ is the number of foreground pixels and mm_per_pixel is the physical pixel size in millimeters. Because per-acquisition geometric calibration (e.g., DICOM PixelSpacing or in-image fiducials) was not available, areas are reported in pixels ($N_{pixels}$ on the 512 × 512 input) rather than in $mm^2$.

**Area Error Metric.** Because areas are measured in pixels without per-acquisition calibration, absolute area errors (e.g., MAE and RMSE) do not reflect true physical measurement error and are not reported.

Instead, prediction accuracy relative to ground truth was quantified using the Mean Absolute Percentage Error (MAPE), which is invariant to the pixel spacing:

$$MAPE = \frac{1}{N} \sum_{i=1}^{N} \frac{|A_{\mathrm{pred},i} - A_{\mathrm{GT},i}|}{A_{\mathrm{GT},i}} \times 100\% \tag{7}$$

In clinical deployment, per-acquisition DICOM PixelSpacing (or an in-image fiducial) converts the predicted pixel area to $mm^2$ through a single multiplicative constant; all MAPE results reported here are invariant to this factor.

**Agreement Analysis.** To assess agreement between predicted and reference areas at the level of the patient, areas were averaged over the frames of each patient (1 to 9 frames per patient, mean 3.5), so that frames from the same patient were not treated as independent. Agreement was assessed with a Bland–Altman analysis [50] of the percentage difference, defined as (predicted - reference) divided by the mean of the two values and multiplied by 100, which does not depend on the pixel spacing. The bias was the mean percentage difference, and the 95% limits of agreement were the bias ± 1.96 times the standard deviation of the differences, with confidence intervals from the approximate variance formulas of Bland and Altman. Proportional bias was tested by regressing the percentage difference on the mean area. Agreement was further quantified with the intraclass correlation coefficient for absolute agreement under a two-way random-effects model, ICC(2,1) [51], with its 95% confidence interval from the F distribution [52]. The 95% confidence interval of the frame-level MAPE was obtained by resampling patients with replacement (5,000 bootstrap samples). All areas were analyzed in pixels, and these statistics are invariant to the pixel spacing.

### *2.4.4 Primary Endpoint*

Among these metrics, area measurement accuracy (MAPE) was treated as the primary evaluation endpoint, reflecting the clinical objective established in Section 1: $PA_{rest}$ functions as a biomarker through its measured area, not through pixel-level boundary tracing. Segmentation overlap (Dice, Jaccard) is reported as a secondary metric, providing a strong technical indicator of model quality independent of clinical framing. Boundary distance metrics (HD95, ASD) are reported as tertiary indicators, given their higher sensitivity to isolated outlier errors relative to overlap-based metrics.

### *2.4.5 Evaluation Protocol*

All metrics were computed on predictions resized to 512 × 512 pixels to ensure fair comparison across models trained on this resolution. For SAM and MedSAM, predictions were generated at the model's native input resolution of 1024 × 1024 pixels using pre-trained weights without fine-tuning (zero-shot setting), and were subsequently resized to 512 × 512 pixels for consistent metric computation. Ground-truth bounding box prompts were provided to SAM and MedSAM during inference to ensure a fair evaluation of their segmentation capability while avoiding the additional challenge of prompt engineering.

# 3. Results

## 3.1 Area Measurement Accuracy

As established in Section 2.4.4, clinical area measurement accuracy is the primary evaluation endpoint for this task. Table 1 presents the quantitative results. GAG-Net achieved the lowest MAPE among all evaluated methods (8.17%), outperforming the next-best method, SAMUS, by 1.90 percentage points despite SAMUS using 3.4× more parameters. Zero-shot foundation models (SAM, MedSAM) exhibited substantially higher area error (>35%), reflecting their reliance on generic visual boundaries rather than the anatomical rules that define PArest.

**Table 1** Quantitative comparison of area measurement accuracy (Mean, Std, MAPE) across state-of-the-art methods. Best result is highlighted in bold. All area measurements are in pixels (areas of the 512 × 512 predictions)

| **Method** | **Mean (px)** | **Std (px)** | **MAPE (%)** |
|---|---|---|---|
| **Ground Truth** | 3663 | 1206 | 0.00 |
| **GAG-Net (Ours)** | 3690 | 1113 | **8.17** |
| U-Net | 3537 | 1227 | 10.66 |
| TransUNet | 3470 | 1326 | 13.30 |
| SwinUNet | 3598 | 1185 | 18.30 |
| SAM (Zero-shot) | 4369 | 1470 | 35.21 |
| MedSAM (Zero-shot) | 4680 | 1585 | 39.18 |
| SAMUS | 3544 | 1145 | 10.07 |

At the patient level (67 patients), the ICC(2,1) between predicted and reference areas was 0.966 (95% CI 0.954–0.977), which is in the excellent range [53]. In the Bland–Altman analysis (Fig. 5b), the mean percentage difference was +1.2% (95% CI -0.9 to +3.3%), which does not indicate systematic over- or underestimation, and the 95% limits of agreement were -15.8% to +18.2%. The difference did not depend on the area (slope -0.001 percentage points per pixel, $p = 0.21$), and 64 of 67 patients (95.5%) lay within the limits. The 95% bootstrap confidence interval of the frame-level MAPE of 8.17% was 6.96–9.51%. The two largest patient-level differences were overestimations of about 20% (Fig. 5b).

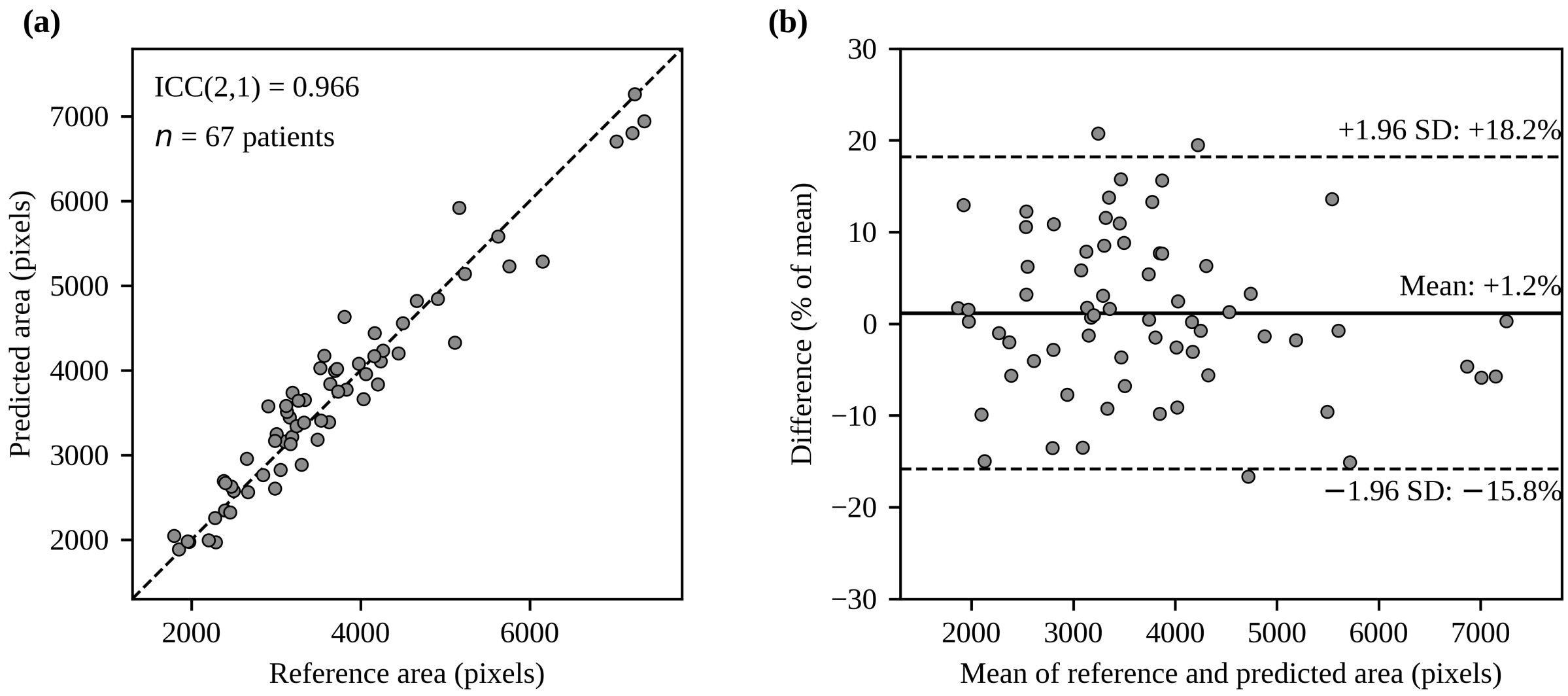

**Fig. 5** Patient-level agreement between predicted and reference $PA_{rest}$ areas (67 patients). (a) Predicted versus reference area; the dashed line is the line of identity. (b) Bland–Altman plot of the percentage difference; the solid line is the mean difference (bias) and the dashed lines are the 95% limits of agreement

## 3.2 Segmentation Overlap

Table 2 reports segmentation overlap metrics. GAG-Net achieved the highest Dice score (0.8862) among all comparisons, outperforming U-Net and TransUNet by 3.25% and 4.40%, respectively, and remained superior to SAMUS (0.8492) despite the latter's substantially larger parameter count. Zero-shot foundation models again underperformed markedly (Dice <0.60), confirming that generic visual prompts are insufficient for the $PA_{rest}$ task without explicit anatomical reasoning.

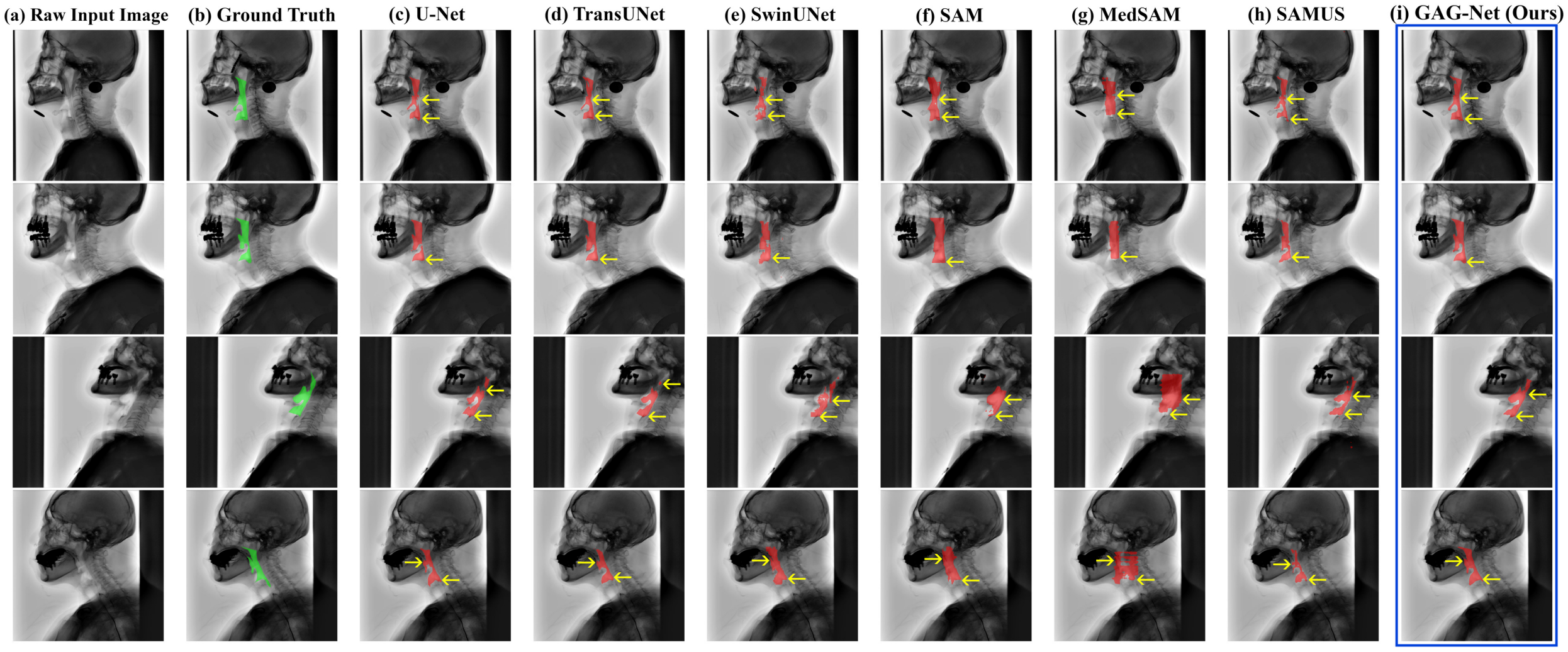

**Fig. 6** Qualitative comparison of segmentation results on the VFSS test set. (a) Raw input image. (b) Ground Truth contours. (c) U-Net [20]. (d) TransUNet [54]. (e) SwinUNet [55]. (f) SAM [47]. (g) MedSAM [48]. (h) SAMUS [49]. (i) GAG-Net (Ours). Contours in (c)–(i) indicate the predicted boundaries. Arrows highlight challenging regions with indistinct soft-tissue boundaries (e.g., pyriform sinus), where competing methods often exhibit

boundary leakage or under-segmentation. In contrast, GAG-Net maintains high anatomical consistency, closely matching the ground truth

To visually corroborate these quantitative findings, Fig. 6 presents a qualitative comparison of segmentation results. As highlighted by the yellow arrows, competing methods (e.g., U-Net and MedSAM) struggle with vague boundaries in the pyriform sinus region, producing leakage or shape distortion. In contrast, GAG-Net generates structurally coherent predictions that align closely with ground truth, demonstrating how anatomical guidance preserves global pharyngeal morphology despite pixel-level boundary variations.

**Table 2** Quantitative comparison of segmentation overlap and boundary distance against state-of-the-art methods. Best results for each metric are highlighted in bold. All distance metrics (HD95, ASD) are reported in pixel units

| Method | Dice | Jaccard | HD95 (px) | ASD (px) | Params (M) |
|---|---|---|---|---|---|
| **GAG-Net (Ours)** | **0.8862** | **0.7986** | 15.84±9.55 | 3.33±1.74 | 39.7 |
| U-Net | 0.8583 | 0.7561 | 16.47±24.19 | 2.28±1.09 | 31.0 |
| TransUNet | 0.8489 | 0.7434 | 17.21±27.58 | 2.55±1.56 | 145.8 |
| SwinUNet | 0.7524 | 0.6091 | 31.86±40.25 | 4.06±1.38 | 27.1 |
| SAM (Zero-shot) | 0.6760 | 0.5352 | 24.99±19.91 | 7.12±7.94 | 93.5 |
| MedSAM (Zero-shot) | 0.5608 | 0.4042 | 25.67±12.19 | 7.97±3.32 | 93.5 |
| SAMUS | 0.8492 | 0.7409 | **9.70**±5.29 | **2.60**±1.03 | 136.3 |

### 3.3 Boundary Distance and Comparison with JOINED

To further validate the structural efficacy and operational efficiency of GAG-Net, we benchmarked it against JOINED [21], a representative two-stage landmark-based segmentation method where a coarse stage (101.9M parameters) generates initial predictions that sequentially inform a fine refinement stage (19.3M parameters). Table 3 presents this comparison alongside boundary distance results.

JOINED exhibits stage-wise performance variation: the coarse stage achieves Dice 0.8856 and HD95 9.19px, while the fine stage yields Dice 0.8839 and HD95 18.52px, suggesting error propagation across its sequential refinement. GAG-Net matches JOINED-Coarse's Dice score (0.8862 vs. 0.8856) within a single end-to-end model, using only 39.7M parameters against JOINED's combined 121.2M.

On boundary distance, GAG-Net's HD95 (15.84px) exceeds both JOINED-Coarse (9.19px) and SAMUS (9.70px, Table 2). As a tertiary metric (Section 2.4.4), HD95 is more sensitive to isolated outlier errors than overlap-based measures, and a gap of about 6 pixels at the 512 × 512 input resolution does not by itself indicate a clinically relevant difference in the measured area: GAG-Net had a lower area error than SAMUS despite its higher HD95 (Tables 1 and 2).

These boundary-level trade-offs stand in contrast to GAG-Net's efficiency profile. With only ∼39.7M parameters, GAG-Net is substantially more compact than TransUNet (∼146M), MedSAM (∼93M),

SAMUS (∼136M, 3.4× heavier), and JOINED's combined total (121.2M), using 67.2% fewer parameters and 73.7% less training VRAM (20GB vs. 76GB for the Coarse stage). Taken together with its leading area measurement accuracy and competitive Dice score, these results indicate that GAG-Net achieves its accuracy through a substantially more efficient architecture, at the cost of secondary boundary-distance precision.

A single end-to-end model also simplifies clinical use compared with a two-stage pipeline. Because JOINED-Fine depends on the outputs of JOINED-Coarse, updating the pipeline on new data requires retraining both stages in sequence, and the larger stage determines the GPU memory required. We did not benchmark inference time, so this argument rests on the architecture and on the training-time measurements in Table 3, not on a measured deployment comparison.

**Table 3** Performance and efficiency comparison with the sequential multi-stage JOINED framework [21]. JOINED-Fine takes the output of JOINED-Coarse as input and cannot run on its own, so the complete pipeline requires both stages (121.2M parameters in total). VRAM is the GPU memory measured during training. In contrast, GAG-Net demonstrates superior accuracy-efficiency balance compared to JOINED in end-to-end training

| Metric | GAG-Net | JOINED-Coarse | JOINED-Fine |
|---|---|---|---|
| Dice | **0.8862** | 0.8856 | 0.8839 |
| HD95 (px) | 15.84 | 9.19 | 18.52 |
| Params (M) | 39.7 | 101.9 | 19.3 |
| Training VRAM (GB) | 20 | 76 | 8 |

## 3.4 Ablation Study

To quantify the specific contributions of the Deformable Convolutional Network (DCN) and the Anatomical Graph Network (AGN), we conducted a comprehensive ablation study. Table 4 summarizes the segmentation performance across four configurations, highlighting the incremental gains provided by each component.

**Table 4** Ablation study on the contributions of DCN and AGN to segmentation performance

| Configuration | Dice | HD95 (px) |
|---|---|---|
| Baseline (no DCN, no AGN) | 0.8390 | **10.92** |
| + DCN only | 0.8526 | 11.65 |
| + AGN only | 0.8843 | 20.14 |
| **+ DCN + AGN (GAG-Net)** | **0.8862** | 15.84 |

**Effect of Deformable Convolution.** We integrated DCN into the encoder to explicitly handle non-rigid pharyngeal deformations. As demonstrated in Table 4, DCN consistently enhanced performance. When added to the baseline (without AGN), DCN improved the Dice score by 1.36%p (0.8390 → 0.8526). Even in the presence of AGN, introducing DCN yielded a further refinement of 0.19%p (0.8843 → 0.8862). These results confirm that DCN effectively captures subtle, fine-grained boundary variations and geometric irregularities that fixed-grid convolutions struggle to resolve, thereby complementing the global structural guidance provided by AGN.

**Effect of AGN.** AGN introduces global anatomical priors to handle morphological variability. The integration of AGN resulted in the most significant performance improvements. Without DCN, adding AGN increased the Dice score by 4.53%p (0.8390 → 0.8843). With DCN, AGN further boosted performance by 3.36%p (0.8526 → 0.8862). These findings indicate that structural guidance is the dominant factor in segmenting the indistinct $PA_{rest}$ boundaries.

Critically, a marginal increase in HD95 was observed upon adding AGN (10.92 px → 15.84 px, +4.9 px). This trade-off is consistent with choosing the measured area as the primary endpoint (Section 2.4.4): HD95 is sensitive to isolated outlier pixels, whereas the overlap and area metrics reflect the quantity used as the biomarker.

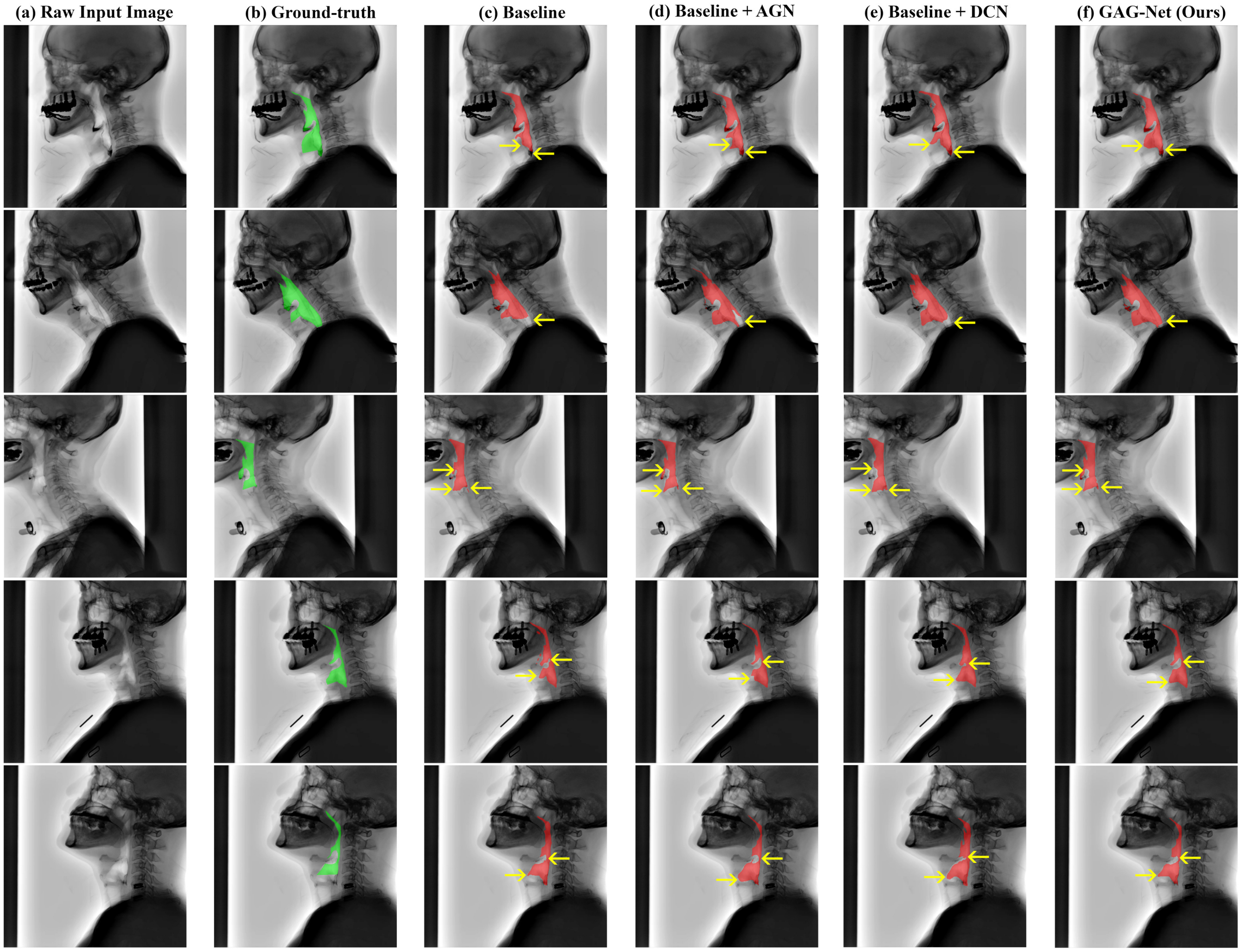

**Fig. 7** Qualitative visualization of the ablation study. (a) Raw Input image. (b) Ground Truth. (c) Baseline (ResNet-50 + U-Net). (d) Baseline + AGN. (e) Baseline + DCN. (f) GAG-Net (Baseline + DCN + AGN). AGN alone (d)

more accurately traces landmark-defined straight-line boundaries and correctly masks the pyriform sinus, while DCN alone (e) yields finer-grained local boundary detail without full adherence to these rules. The full model (f) combines both properties, achieving the closest agreement with the ground truth

The qualitative improvements driven by each component are visualized in Fig. 7. Adding AGN alone (Fig. 7(d)) yields two distinct improvements over the baseline: it correctly masks the pyriform sinus despite residue obscuring part of the region (right arrow), and it more accurately traces the straight-line segments prescribed by the annotation protocol — for instance, the boundary connecting the nasal spine and PW (beginning of the posterior pharyngeal wall), and the perpendicular segment extending from the top of the arytenoid cartilage to the anterior pharyngeal wall (left arrow). This indicates that AGN successfully learns the geometric constraints linking specific landmarks, consistent with its role in encoding relational structure. Adding DCN alone (Fig. 7(e)) shows less precise adherence to these landmark-defined rules but produces finer-grained boundary detail closer to the ground truth than the baseline, reflecting its strength in local boundary adaptation. The full model (Fig. 7(f)) combines both properties, yielding the closest overall agreement with the ground truth and confirming that global geometric guidance (AGN) and local boundary refinement (DCN) are complementary.

**Summary.** The ablation results confirm the complementary roles of the proposed components: DCN enhances local adaptability to boundary deformations, while AGN ensures global anatomical coherence. The substantial gain from AGN specifically highlights that $PA_{rest}$ segmentation relies more heavily on anatomical context than on local texture features alone. Notably, the deliberate trade-off, sacrificing marginal HD95 precision for superior Dice and area accuracy, aligns with clinical priorities where volumetric and structural fidelity outweigh pixel-level boundary exactitude.

# 4. Discussion

## 4.1 Anatomical Priors in $PA_{rest}$ Segmentation

The primary finding of this study is that explicitly embedding anatomical landmark information into the segmentation pipeline substantially improves performance on the $PA_{rest}$ task. Unlike conventional segmentation targets with visually distinct boundaries, $PA_{rest}$ delineation follows clinician-defined rules based on landmark connectivity [9, 10], meaning that the segmentation problem is inherently constrained by structural knowledge that pixel-level learning alone cannot capture. By introducing landmark coordinates through the GAGM, GAG-Net effectively constrains the segmentation output space, reducing the degrees of freedom available to the decoder and guiding predictions toward anatomically plausible configurations.

This finding is consistent with the broader literature on anatomy-aided segmentation. Liu et al. [28] identified graph neural network-based structural reasoning as a promising but underexplored direction for incorporating anatomical priors. Our work provides empirical evidence supporting this direction: the ablation study (Table 4) shows that the AGN contributes a 4.53%p Dice improvement over the baseline, substantially exceeding the 1.36%p gain from DCN alone. This confirms that for targets where boundaries are clinically defined rather than visually apparent, global structural context is more critical than local boundary adaptation.

Prior approaches to anatomical priors have relied on statistical atlases [29, 30] or post-processing corrections [34], which assume relatively stable organ morphology. The pharyngeal region, however, exhibits substantial inter-patient variability in length, shape, and laryngeal positioning due to its composition of diverse muscles and cartilages involved in swallowing [11]. Atlas-based methods struggle to accommodate such variability because they rely on population-level shape templates. GAG-Net addresses this by generating instance-specific structural guidance: the AGN constructs a guidance map from landmarks predicted for each individual image, adapting to patient-specific anatomy rather than enforcing a fixed prior.

## 4.2 Role of Deformable Convolution

The complementary role of DCN is to handle the non-rigid, irregular contours characteristic of the pharyngeal region. The $PA_{rest}$ boundary follows complex polygon-like shapes formed by soft tissues, rather than smooth elliptical or rectangular contours typical of other segmentation targets. Standard convolutions with fixed sampling grids cannot adapt to such geometries. The ablation study confirms that DCN consistently improves performance (+1.36%p without AGN, +0.19%p with AGN), indicating that adaptive local feature extraction complements global anatomical reasoning.

The relatively smaller marginal gain of DCN when AGN is present (+0.19%p vs. +1.36%p) suggests that AGN already captures much of the structural information that DCN provides independently. Nevertheless, DCN remains valuable for capturing fine-grained boundary variations that graph-level reasoning cannot resolve, as confirmed by the qualitative results in Fig. 7. This observation is consistent with Hu et al. [40], who demonstrated that adaptive convolution strategies improve segmentation of morphologically heterogeneous anatomical structures.

## 4.3 Foundation Models for Domain-Specific Segmentation

The poor zero-shot performance of SAM [47] (Dice: 0.6760) and MedSAM [48] (Dice: 0.5608) highlights a fundamental limitation of vision foundation models for this task. These models are trained to segment objects along visually apparent boundaries, whereas $PA_{rest}$ boundaries are defined by clinical guidelines that reference anatomical landmarks rather than intensity gradients [9, 18, 19]. Without knowledge of these clinical rules, foundation models cannot infer where the boundary should be, regardless of their capacity for general visual understanding. This observation aligns with recent comprehensive reviews [25, 26] that emphasize the need for specialized architectural designs in automated VFSS analysis.

Even with domain-specific fine-tuning, SAMUS [49] achieved a lower Dice score (0.8492) than GAG-Net despite using 3.4× more parameters (136.3M vs. 39.7M). SAMUS achieved the lowest HD95 (9.70 px) through its dual CNN-Transformer architecture that captures both local and global features. One possible explanation is that SAMUS produces morphologically stable but spatially biased predictions. Its pre-trained SAM branch likely provides strong shape regularization, reducing boundary outliers; however, without explicit landmark guidance, the predicted region may be systematically offset from the clinically defined $PA_{rest}$.

## 4.4 Clinical Significance of Area Measurement Accuracy

From a clinical perspective, a key finding is GAG-Net's low area measurement error (MAPE: 8.17%). Miyashita et al. [10] established that $PA_{rest}$ measurement is significantly associated with the presence of sarcopenic dysphagia ($p = 0.02$), and Molfenter et al. [11] confirmed through CT-based volumetry that increased pharyngeal area reflects muscular degeneration. For such quantitative biomarkers, area measurement reliability is more clinically meaningful than pixel-level boundary precision. At the patient level, agreement with the reference annotation was excellent (ICC 0.966), and the limits of agreement of about ±17% indicate that an individual measurement can still deviate noticeably from the reference, which matters when the area is compared with a diagnostic threshold. These statistics describe agreement with a single reference annotation, not agreement between independent raters.

The comparison between GAG-Net and SAMUS exemplifies this distinction: despite SAMUS achieving lower HD95 (9.70 vs. 15.84 px), GAG-Net shows lower area measurement error (8.17% vs. 10.07%). HD95 is determined by the worst 5% of boundary points, whereas the area error reflects the whole enclosed region, so the two can rank methods differently.

## 4.5 Limitations and Future Work

Several limitations should be acknowledged. First, this study relies on a single-center dataset of 2,255 frames from 137 patients at one institution. While the patient-level data split mitigates within-patient bias, generalizability to VFSS images acquired with different fluoroscopy equipment, imaging protocols, or patient populations remains to be validated. Multi-center studies are needed to assess robustness across clinical settings.

Second, all annotations were performed by a single trained annotator verified by a senior specialist. Although this ensures high annotation consistency, the absence of multiple independent annotators precludes formal inter-rater reliability assessment. Future work should incorporate annotations from multiple clinicians to quantify inter-rater agreement and establish measurement uncertainty bounds, and should add external validation, as done for keypoint-based Insall–Salvati ratio measurement [17].

Third, the current AGN employs a standard GCN with a fixed chain adjacency structure. More expressive graph architectures that learn edge weights or update edge representations during message passing could potentially capture richer anatomical dependencies. Exploring such alternatives is a natural direction for future investigation.

Fourth, GAG-Net processes individual frames independently without leveraging temporal information from VFSS video sequences. Incorporating temporal modeling could improve segmentation consistency across consecutive frames and enable frame-to-frame annotation propagation, substantially reducing the annotation burden for future dataset construction. Relatedly, rest frames were selected manually in this study, and automatic detection of the rest frame from the video remains to be developed.

Fifth, the mechanisms underlying the trade-off between overlap-based metrics (Dice) and boundary-distance metrics (HD95) when DCN and AGN are combined remain incompletely understood, and disentangling their distinct effects on boundary precision is a priority for future work.

Finally, while the GAGM is designed as a modular component, its portability to other backbone architectures (e.g., Transformer-based encoders such as SAM) has not been evaluated. Validating the GAGM as a transferable module across different segmentation frameworks, as well as extending it to other anatomically complex structures beyond the pharyngeal region, are promising directions for future work.

## 5. Conclusions

This study presented GAG-Net, an automated $PA_{rest}$ measurement framework for rest frames that integrates graph-based anatomical reasoning with adaptive deformable convolutions to measure $PA_{rest}$ in VFSS images, together with an explicit annotation protocol and a specialist-verified dataset of 2,255 frames from 137 patients. On the patient-level held-out test set, GAG-Net achieved the lowest clinical area measurement error (MAPE: 8.17%), the primary endpoint, together with the highest segmentation accuracy (Dice: 0.8862) among all evaluated methods, with only 39.7M parameters. At the patient level, agreement with the reference annotation was excellent (ICC 0.966; 95% limits of agreement -15.8% to +18.2%). The ablation study showed that AGN produced a larger improvement in Dice than DCN alone, supporting the role of explicit anatomical guidance for this clinically defined target. The proposed framework provides a standardized, computationally efficient tool for quantitative assessment of sarcopenic dysphagia, potentially reducing reliance on costly supplementary imaging modalities. Future work will focus on multi-center validation, calibrated absolute-area measurement and prospective clinical evaluation, temporal extension to video-level analysis, and evaluation of the GAGM as a transferable module for other anatomically complex segmentation tasks.

## Acknowledgements

The authors would like to thank the members of our research laboratory for their insightful discussions and technical assistance.

## Statements and Declarations

**Funding.** The present research was conducted by the research fund of Dankook University in 2026.

**Competing interests.** The authors declare that they have no conflict of interest.

**Ethics approval and consent to participate.** This retrospective study involving human participants was in accordance with the ethical standards of the institutional and/or national research committee and with the 1964 Helsinki Declaration and its later amendments or comparable ethical standards. The study protocol was approved by the Institutional Review Board of Chung-Ang University Hospital (IRB No. 2501-007-19557). Informed consent was waived due to the retrospective nature of the study and the use of de-identified clinical data.

**Data and code availability.** The data used in this study are not publicly available due to privacy restrictions on patient clinical data (VFSS videos). The data are available from the corresponding author on reasonable request, subject to institutional IRB approval. The source code will be made publicly available upon acceptance.

**Author contributions.** JunYeong Park: Conceptualization of this study, Methodology, Software, Formal analysis, Investigation, Visualization, Data curation, Writing – Original Draft, Writing – Review & Editing. Myung Woo Park: Conceptualization of this study, Supervision, Methodology, Data curation, Writing – Review & Editing. Byung Chan Lee: Resources, Data curation. Hyung Jin Chang: Supervision, Writing – Review & Editing. Sang-Il Choi: Conceptualization of this study, Formal analysis, Supervision, Project administration, Funding acquisition, Resources, Writing – Review & Editing.